\documentclass[conference]{IEEEtran}

\usepackage[numbers,sort&compress]{natbib}
\usepackage{times}
\usepackage{multirow}
\usepackage{xcolor}
\usepackage{color}
\definecolor{Brown}{cmyk}{0, 0.8, 1, 0.6}

\newcommand\bm[1]{\mathbf{#1}}
\usepackage{booktabs}

\usepackage{lscape}
\usepackage{color}
\usepackage{amssymb}
\usepackage{amsmath}
\usepackage{balance}  
\usepackage{graphicx} 
\usepackage{times}    
\usepackage{url}      
\usepackage[lofdepth,lotdepth]{subfig}
\usepackage{multirow}
\usepackage{bbm}
\usepackage{pseudocode}
\usepackage{color}
\usepackage{amsmath}
\usepackage{algorithm}
\usepackage[noend]{algpseudocode}

\newtheorem{definition}{Definition}

\newtheorem{assumption}{Assumption}

\newcommand{\TheName}{\texttt{\bf DTNH}}
\newcommand{\vc}[1]{\mathbf{#1}}

\begin{document}

\title{Towards Making Deep Transfer Learning\\ Never Hurt}
\author{Ruosi Wan$^{1,3,\dag}$, Haoyi Xiong$^{1,2,\dag,*}$, Xingjian Li$^{1,2}$, Zhanxing Zhu$^3$ and Jun Huan$^{1,2}$\\
$^1$ Big Data Laboratory, Baidu Inc., Beijing, China\\
$^2$ National Engineering Laboratory for Deep Learning Technology and Applications, Beijing, China\\
$^3$ School of Mathematical Sciences, Peking University, Beijing, China}
\maketitle

\begin{abstract}
Transfer learning have been frequently used to improve deep neural network training through incorporating weights of pre-trained networks as the starting-point of optimization for regularization. While deep transfer learning can usually boost the performance with better accuracy and faster convergence, transferring weights from inappropriate networks hurts training procedure and may lead to even lower accuracy. 
In this paper, we consider deep transfer learning as minimizing a linear combination of empirical loss and regularizer based on pre-trained weights, where the regularizer would restrict the training procedure from lowering the empirical loss, with conflicted descent directions (e.g., derivatives).

Following the view, we propose a novel strategy making regularization-based Deep Transfer learning Never Hurt (\TheName) that, for each iteration of training procedure, computes the derivatives of the two terms separately, then re-estimates a new descent direction that does not hurt the empirical loss minimization while preserving the regularization affects from the pre-trained weights. Extensive experiments have been done using common transfer learning regularizers, such as $L^2$-SP and knowledge distillation, on top of a wide range of deep transfer learning benchmarks including Caltech, MIT indoor 67, CIFAR-10 and ImageNet. The empirical results show that the proposed descent direction estimation strategy \TheName\ can always improve the performance of deep transfer learning tasks based on all above regularizers, even when transferring pre-trained weights from inappropriate networks. All in all, \TheName\ strategy can improve state-of-the-art regularizers in all cases with 0.1\%---7\% higher accuracy in all experiments.
\end{abstract}

\section{Introduction}
In real-world applications, the performance of deep learning often is limited by the size of training set. Training a deep neural network with a small number of training instances usually results in the so-called \emph{over-fitting} problem and the generalization capability of the obtained model is low. A simple yet effective approach to obtain high-quality deep neural network models is transfer learning~\cite{pan2010survey} from pre-trained models. In such practices~\cite{donahue2014decaf}, a deep neural network is first trained using a large (and  possibly irrelevant) source dataset (e.g. ImageNet). The weights of such a network are then fine-tuned using the data from the target application domain.  

Through incorporating the weights of appropriate pre-trained networks as starting-points of optimization and/or reference for regularization~\cite{li2018explicit,yim2017gift}, deep transfer learning can usually boost the performance with better accuracy and faster convergence. For example, \emph{Li et al.}~\cite{li2018explicit} recently proposed $L^2$-SP algorithm that leverages the squared Euclidean distance between the weights of source and target networks as a regularizer for knowledge transfer and further set the weights of source network as the starting point for optimization procedure. In this approach, $L^2$-SP transfers the ``knowledge'' from the pre-trained source networks to the target ones, through constraining the difference of weights between the two networks, while minimizing the empirical loss on the target task. In addition to the direct regularization on weights, ~\cite{yim2017gift} demonstrated the capacity of \emph{``knowledge distillation''}~\cite{hinton2015distilling} to transfer knowledge from the source to target networks in an teacher-student training manner, where the source network performs as a teacher regularizing the outputs from the outer-layers of target network. All in all, knowledge distillation based approach intends to transfer the capacity of feature learning from the source network to the target one, through constraining the difference of feature maps outputted by the outer layers (e.g., convolution layers)~\cite{yosinski2014transferable,huh2016makes} of the two networks.

In addition to aforementioned strategies, a great number of methods, e.g.,~\cite{kirkpatrick2017overcoming,li2017learning}, have been proposed to transfer the knowledge from the weights of pre-trained source networks to the target task, for better accuracy. However, incorporating weights from inappropriate networks using inappropriate transfer learning strategies sometimes may hurt the training procedure and may lead to even lower accuracy. This phenomena is called ``negative transfer''~\cite{rosenstein2005transfer,pan2010survey}. For example, ~\cite{yosinski2014transferable} observed that reusing the pre-trained weights of ImageNet task through inappropriate ways could poison the CNN training and forbids deep transfer learning achieving its best performance. It indicates that reusing pre-trained weights from inappropriate datasets and/or via inappropriate transfer learning strategies hurts deep learning.
\begin{figure}
    \centering
    \includegraphics[width=0.5\textwidth]{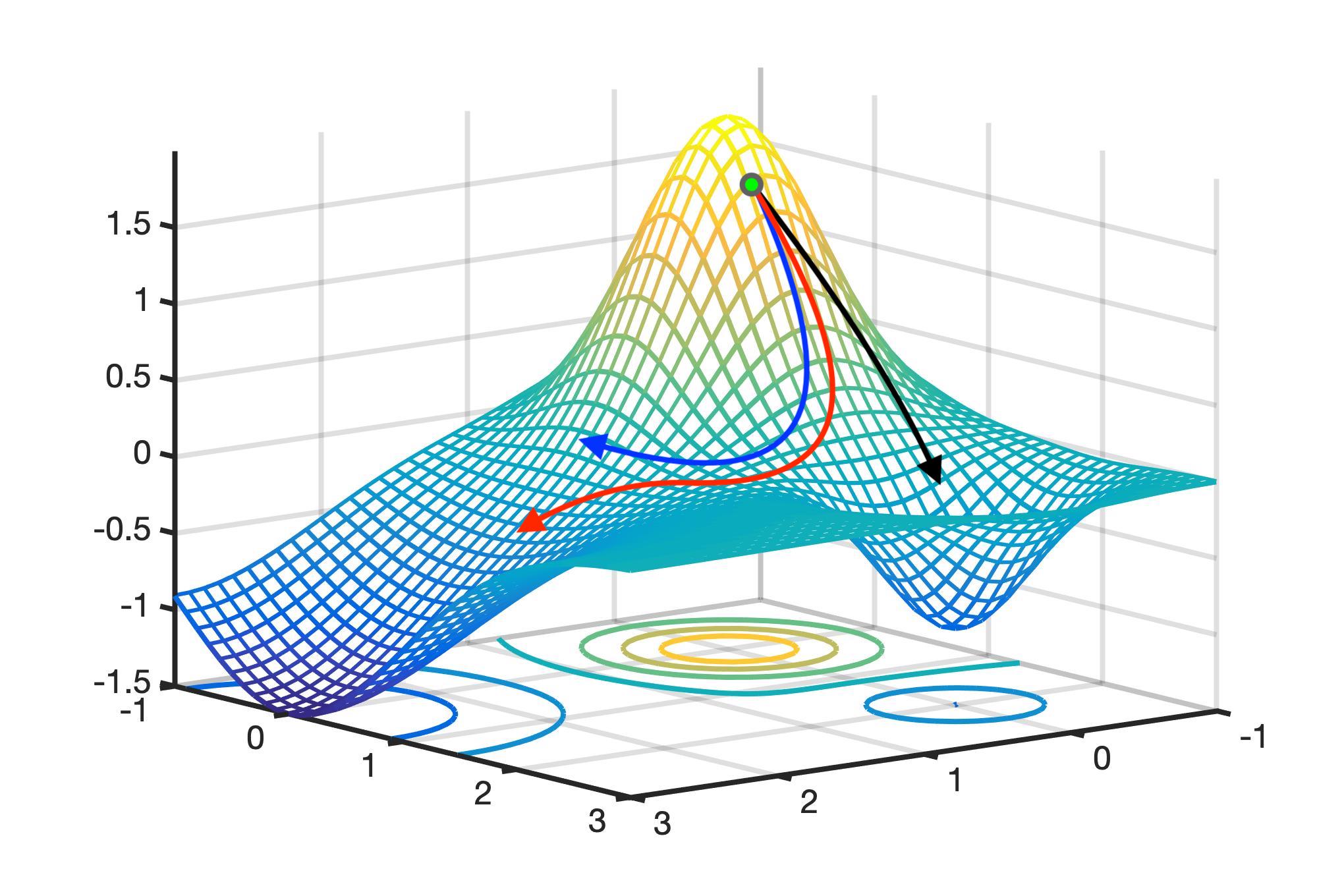}
    \caption{Flows of Descent Directions on the Empirical Loss. \textcolor{black}{\bf Black Line}: the flow via descent direction of empirical loss (i.e., gradient of empirical loss) from pre-trained weights as the starting point, where the descent direction quickly leads the learning procedure converged to a local minimum of over-fitting; \textcolor{blue}{\bf Blue Line}: the flow via the descent direction linearly combining gradients of empirical loss and $L^2$-SP regularizer~\cite{li2018explicit} from pre-trained weights as the starting point, where the regularization hurts the minimization of empirical loss; \textcolor{red}{\bf Red Line}: the flow via the descent direction balancing the gradients of empirical loss and $L^2$-SP regularizer, where the descend direction leads to a flat area with low empirical loss (i.e., potentials of great generalizability).}
    \label{fig:descent}
\end{figure}

In this paper, we consider deep transfer learning as minimizing a linear combination of empirical loss and regularizer based on pre-trained weights, where the empirical loss refers to fitness on the target dataset and regularization controls the divergence (of weights, feature maps, and etc.) between source and target networks.
From an optimization perspective,  the regularizer would restrict the training procedure from lowering the empirical loss, as the descent direction (e.g., derivatives) of the regularizer might be conflicted with the descent direction of empirical loss. 

We specify above observation using an example based on $L^2$-SP~\cite{li2018explicit} shown in Figure~\ref{fig:descent}. The Black Line refers to the empirical loss descent flow of common gradient-based learning algorithms with pre-trained weights as the start point. It shows that with the empirical loss gradients as the descent direction, such method quickly converges to a local minimum in a narrow cone, which is usually considered as an over-fitting solution. In the meanwhile, the Blue line demonstrates the possible empirical loss descending path of $L^2$-SP algorithm, where a strong regularization blocks the learning algorithm to continue lowering the empirical loss while traversing the area around the point of pre-trained weights. An ideal case has been illustrated as the red line, where $L^2$-SP regularizer helps the learning algorithm to avoid the over-fitting solutions. The overall descent direction adapting $L^2$-SP regularizer with respect to empirical loss leads to generalizable solutions. There thus needs a method to make both empirical loss and regularizer continue descending to boost the performance of deep transfer learning.


\textbf{Our Contribution.} In this way, we propose a novel deep transfer learning strategy \texttt{DTNH} that makes regularization-based Deep Transfer learning Never Hurt.
Specifically, for each iteration of training procedure, DTNH computes the derivatives of the empirical loss and regularizer terms separately, then re-estimates a new descent direction that doesn't hurt the empirical loss minimization while preserving the regularization affects from the pre-trained weights. Extensive experiments have been done using common transfer learning regularizers, such as $L^2$-SP and knowledge distillation, on top of a wide range of deep transfer learning benchmarks including Caltech, MIT indoor 67, CIFAR-10 and ImageNet. The experiments shows that \texttt{DTNH} can always improve the performance of deep transfer learning tasks, even when reusing pre-trained weights from inappropriate networks (i.e., the performance of vanilla tranfer learning from the source task is even worse than direct training using the target dataset only). All in all, \TheName\ can work with above regularizers in all cases with 0.1\% -- 7\% higher accuracy than state of the art algorithms.

\textbf{The Organization of the paper.} The rest of this paper is organized as follow. In Section II, we present the introduction to the relevant work of this research and discuss the comparisons to our work. Section III briefs the preliminary work and the technical backgrounds of our work, where the state of the art models for the regularization-based deep transfer learning are introduced with details. Section IV presents the design of the proposed algorithm~\TheName, where we first formulate the research problem of this work with two key assumptions, then present the algorithm design and discuss the novelty of the proposed algorithms.  In Section V, we majorly report the experiments that we conducted to validate \TheName, where we first present the experiment setups and datasets used, then introduce the main results with the accuracy comparisons between \TheName\ and the existing transfer learning algorithms, further we provide several case studies that validate the two key assumptions made for problem formulation. Finally, we discuss the open issues of this work in Section VI and conclude this work in Section VII.

\section{Related Work}
In this section, we introduce the related work of deep transfer learning with the most relevant work to our study emphasized, where we first introduce the existing work in deep transfer learning in general, then emphasize the deep transfer learning algorithms that uses pre-trained models for the regularization, further we discuss the connection of existing work to this paper.

%
\subsection{Transfer Learning with Deep Architectures}
Transfer learning refers to a type of machine learning paradigms that aim at transferring knowledge obtained in the source task to a (maybe irrelevant) target task~\cite{pan2010survey,caruana1997multitask}, where the source and target tasks can share either same or different label spaces. In our research, we primarily consider the inductive transfer learning with different target label space for deep neural networks. 
As early as in 2014, authors in ~\cite{donahue2014decaf} reported their observation of significant performance improvement through directly reusing weights of the pre-trained source network to the targt task, when training a large CNN with tremendous number of filters and parameters. However, in the meanwhile of reusing all pre-trained weights, the target network might be overloaded by learning tons of inappropriate features (that cannot be used for classification in the target task), while the key features of the target task have been probably ignored. In this way, Yosinki \emph{et al.}~\cite{yosinski2014transferable} proposed to understand whether a feature can be transffered to the target network, through quantifying the ``transferability'' of features from each layer considering the performance gain. Furthermore, Huh \emph{et al.}~\cite{huh2016makes} made empirical study on analyzing features that CNN learned from ImageNet dataset to other computer vision tasks, so as to detail the factors effecting deep transfer learning accuracy. In recent days, this line of research has been further developed with increasing number of algorithms and tools that can improve the performance of deep transfer learning, including subset selection ~\cite{ge2017cvpr,cui2018large}, sparse transfer \cite{liu2017sparse}, filter distribution constraining \cite{aygun2017exploiting}, parameter transfer ~\cite{zhang2018parameter}.

\subsection{Regularization-based Deep Transfer Learning}
In our research, we focus on the knowledge transfer through reusing pre-trained weights for the regularization. We categorized the most recent related work as follow. 

\begin{enumerate}

    \item \emph{Regularizing through Weight Distance - } The square Euclidean distance between the weights of source and target networks are frequently used as the regularizer for deep transfer learning~\cite{li2018explicit}.  Specifically, ~\cite{li2018explicit} studied to accelerate deep transfer learning while preventing fine-tuning from over-fitting, using a simple $L^2$-norm regularization on top of the ``Starting Point as a Reference'' optimization. Such method, namely $L^2$-SP, can significantly outperform a wide range of regularization-based deep transfer learning mechanisms, such as the standard $L^2$-norm regularization. 

     \item \emph{Regularizing through Knowledge Distillation - } Yet another way to regularize the deep transfer learning is ``knowledge distillation''~\cite{hinton2015distilling,romero2014fitnets}. In terms of methodologies, the knowledge distillation was originally proposed to compress deep neural networks~ \cite{hinton2015distilling,romero2014fitnets} through teacher-student network training, where the teacher and student networks are usually based on the same task~\cite{hinton2015distilling}. In terms of inductive transfer learning, authors in~\cite{yim2017gift} were first to investigate the possibility of using the distance of intermediate results (e.g., feature maps generated by the same layers) of source and target networks as the regularization term. Further,~\cite{zagoruyko2016paying} proposed to use the distance between activation maps as the regularization term for so-called ``attention transfer''. 

\end{enumerate}
In addition to the use of above two types of regularization, fine-tuning from a pre-trained model with a $L^2$-norm regularization is also appreciated by the deep transfer learning community~\cite{donahue2014decaf}.

\subsection{Discussion on the Connection to Our Work} 
Compared to above work and other transfer learning studies, our work aims at providing a \emph{generic descent direction estimation strategy} that improves the performance of {regularization-based deep transfer learning}. The intuition of \TheName\ is, per iteration during the learning procedure, re-estimating a new direction of loss descending that addresses the affect of regularizers while making the empirical loss reduction/minimization not hurt. In our work, we demonstrated the capacity of \TheName\ working with two most recent deep transfer learning regularizers--$L^2$-SP~\cite{li2018explicit} and Knowledge distillation~\cite{yim2017gift}, which are based on two typical deep learning philosophies (i.e., constraining weights and feature maps respectively), using a wide range of transfer learning tasks. The consistent performance boosts with \TheName\ in all cases of experiments suggests that \TheName\ can improve above regularization-based deep transfer learning with higher accuracy.
%
%
%

Other techniques, including continual learning \cite{kirkpatrick2017overcoming,li2017learning}, attention mechanism for CNN models~\cite{mnih2014recurrent,xu2015show,yang2016stacked,zagoruyko2016paying} and so on, can also improve the performance of knowledge transfer between tasks. We believe our work made complementary contributions in this area. All in all, we appreciate the contributions made by these studies.

\section{Preliminaries and Backgrounds}
In this section, we first introduced the preliminary setting of regularization-based transfer learning, then introducing the backgrounds of $L^2$-SP and knowledge distillation based transfer learning that would be used in our studies.

\subsubsection{Regularization-based Transfer Learning} 
Deep convolutional networks usually consist of a great number of parameters that need fit to the dataset. For example, ResNet-110 has more than one million free parameters. The size of free parameters causes the risk of over-fitting. Regularization-based transfer learning is the technique to reduce this risk by constraining the parameters within a limited space with respect to a set of pre-trained weights. The general learning problem is usually formulated as follow.

\begin{definition}[Regularization-based Deep Transfer Learning]
Let's first denote the dataset for the desired task as $\mathbf{D}=\{(\vc{x}_1,y_1),(\vc{x}_2,y_2),(\vc{x}_3,y_3)\dots,(\vc{x}_n,y_n)\}$, where totally $n$ tuples are offered and each tuple $(\vc{x}_i,y_i)$ refers to the input image and its label in the dataset. 
We then denote $\mathbf{\omega}\in\ \mathbb{R}^{d}$ be the $d$-dimensional parameter vector containing all $d$ parameters of the target model. Further, given a pre-trained network with parameter $\mathbf{\omega_s}$ based on an extremely large dataset as the source, one can estimate the parameter of target network through the transfer learning paradigms. 
The optimization object with regularization-based deep transfer learning is to obtain the minimizer of $\mathcal{L}(\mathbf{\omega})$
\begin{equation} 
\underset{w}{\mathrm{min}}\ \mathcal{L}(\omega)=\left\{ \frac{1}{n}\sum_{i=1}^n L(z( \vc{x}_{i}, \mathbf{\omega}), y_{i}) + \lambda\cdot\Omega(\mathbf{\omega},\mathbf{\omega}_s)\right\} \qquad
\label{eq:deep}
\end{equation}
%
where (i) the first term $\sum_{i=1}^n L(z( \vc{x}_{i}, \mathbf{\omega}), y_{i})$ refers to the empirical loss of data fitting while (ii) the second term $\Omega(\mathbf{\omega},\mathbf{\omega_s})$ characterizes the differences between the parameters of target and source network. The tuning parameter $\lambda >0\ $ balances the trade-off between the empirical loss and the regularization term.
\end{definition}

\subsubsection{Deep Transfer Learning via $L^2$-SP and Knowledge Distillation}
As was mentioned, two common regularization-based deep transfer learning algorithms studied in this paper are $L^2$-SP~\cite{li2018explicit} and knowledge distillation based regularization~\cite{yim2017gift}. Specifically, these two algorithms can be implemented with the general regularization-based deep transfer learning procedure with objective function listed in Eq.~\ref{eq:deep} using following two regularizers
\begin{itemize}
    \item L2-SP \cite{li2018explicit} -- In terms of regularizer, this algorithm uses the squared-euclidean distance between the target weights (i.e., optimization objective $\mathbf{\omega}$) and the pre-trained weights $\mathbf{\omega_s}$ of source network (listed in Eq~\ref{eq:l2sp}) to constrain the learning procedure.
    \begin{equation}
        \Omega(\mathbf{\omega},\mathbf{\omega}_s) = \|\mathbf{\omega} - \mathbf{\omega}_{s}\|_2^2\label{eq:l2sp}
    \end{equation}
    In terms of optimization procedure, $L^2$-SP makes the learning procedure start from the pre-trained weights (i.e., using $\mathbf{\omega_s}$ to initialize the learning procedure).

    \item Knowledge Distillation based Regularization \cite{yim2017gift} --- Given the target dataset $\{(\vc{x}_1,y_1),\dots,(\vc{x}_n,y_n)\}$ and $N$ filters in the target/source networks for knowledge transfer, this algorithm models the regularization as the aggregation of squared-euclidean distances between feature maps outputted by the $N$ filters of the source/target networks, such that
    \begin{equation}
        \Omega(\mathbf{\omega},\mathbf{\omega_s})=\frac{1}{n} \sum_{j=1}^N\sum_{i=1}^n\|\bm{F}_{j}(\mathbf{\omega}, \vc{x}_i) - \bm{F}_{j}(\mathbf{\omega_s}, \vc{x}_i)\|_2^2
    \end{equation}
    where $\bm{F}_j(\mathbf{\omega},\vc{x}_i))$ refers to the feature map outputted by the $j^{th}$ filter ($1\leq j\leq N$) of the target network based on weight $\mathbf{w}$ using input image $\vc{x}_i\ $  ($1\leq i\leq n$). The optimization algorithm can starts from $\mathbf{\omega_s}$ as the initialization of learning.
\end{itemize}
In the rest of this work, we presented a strategy \TheName\ to improve the general form of regularization-based deep transfer learning shown in Eq.~\ref{eq:deep}, then evaluated and compared \TheName\ using above two regularizers with common deep transfer learning benchmarks.

\section{\TheName: Towards Making Deep Transfer Learning Never Hurt}
In this section, we formulate the technical problems of our research with assumptions addressed, then present the design of our solution \TheName.

    \subsection{Problem Formulation}
Prior to formulating our research problem based on the settings, this section introduced the settings and assumptions of the problem.

\begin{definition}[Descent Directions]
Gradient-based learning algorithms are frequently used for deep transfer learning to minimize the loss function listed in Eq.~\ref{eq:deep}. In each iteration of learning procedure, the algorithms estimate a descent direction $\mathbf{d}(\mathbf{\omega})$, such as stochstic gradient, based on the optimization objective $\mathbf{\omega}$ that approximates the gradient, such that
\begin{equation}
\begin{aligned}
  \mathbf{d}(\omega)\approx\nabla \mathcal{L}(\mathbf{\omega})& =  \sum_{i=1}^n \nabla L(z( \vc{x}_{i}, \mathbf{\omega}), y_{i}) + \lambda\nabla\Omega(\mathbf{\omega},\mathbf{\omega}_s)\\
     & = \nabla J(\mathbf{\omega}) +\lambda\cdot\nabla\Omega(\mathbf{\omega},\mathbf{\omega}_s),
\end{aligned}
\end{equation}
where $\nabla J(\mathbf{\omega})= \sum_{i=1}^n \nabla L(z( \vc{x}_{i}, \mathbf{\omega}), y_{i})$ refers to the gradient of empirical loss based on training set and $\nabla\Omega(\mathbf{\omega},\mathbf{\omega}_s)$ is the gradient of regularization term all based on optimization objective $\mathbf{\omega}$. 
\end{definition}

\subsubsection{Key Assumptions}
Due to the affect of regularization $\Omega(\mathbf{\omega},\mathbf{\omega}_s)$, the angle between the actual descent direction $\mathbf{d}(\mathbf{\omega})$ and the gradient of empirical loss $\nabla J(\mathbf{\omega})$, i.e., $\measuredangle (\mathbf{d}(\mathbf{\omega}),\nabla J(\mathbf{\omega}))$, would be large. It is intuitive to state that when $\measuredangle (\mathbf{d}(\mathbf{\omega}),\nabla J(\mathbf{\omega})$ is large, the descent direction cannot effectively lower the empirical loss and causes the potential performance bottleneck of deep transfer learning.
%
We thus formulate the technical problem with following assumptions specified. 

\begin{assumption}[Efficient Empirical Loss Minimization]
 It is reasonable to assume that the descent direction $\mathbf{\widehat d}(\mathbf{\omega})$ having a smaller angle with the gradient of empirical loss, i.e., a smaller $\measuredangle (\mathbf{\widehat d}(\mathbf{\omega}),\nabla J(\mathbf{\omega}))$, can lower the empirical loss more efficiently.
\end{assumption}

\begin{assumption}[Regularization Affect Preservation]
 It is reasonable to assume that there exists a potential threshold $s$ that, when $\measuredangle (\mathbf{\widehat d}(\mathbf{\omega}),\nabla \Omega(\mathbf{\omega},\mathbf{\omega}_s))\leq s$, the descent direction can preserve the affect of the regularization for knowledge transfer.
\end{assumption}

\subsubsection{The Problem} 
Based on above definitions and assumptions, the problem of this research is to propose a new direction descent algorithm---every iteration of the algorithm re-estimates a new descent direction  $\mathbf{\widehat d}$ to lower the training loss based on the optimization object $\mathbf{\omega}$, such that
\begin{equation}\small
    \mathbf{\widehat d(\omega)}\gets\underset{\forall d\in w}{\mathrm{argmin}}\ \measuredangle(d,\nabla J(\omega))\ s.t.\ \measuredangle(d,\nabla \Omega(\mathbf{\omega},\mathbf{\omega_s}))\leq s,
\end{equation}
where $s$ refers to the maximal angle allowed between actual descent direction and the gradient regularizer to preserve the affect of regularizer $\Omega(\mathbf{\omega},\mathbf{\omega_s})$ (\textbf{Assumption 2}). Note that in this research we don't intend to study the exact setting of $s$ and our algorithm implementation is indeed independent with the setting of $s$.

\subsection{\TheName: Descent Direction Estimation Strategy}
In this section, we presented the design of \TheName\ as a descent direction estimator that solves the problem addressed. 
Given the empirical loss function $J(\omega)$, the regularization term $\Omega(\mathbf{\omega},\mathbf{\omega_t})$, the set of training data $\mathbf{D}=\{(\vc{x}_1,y_1),(\vc{x}_2,y_2),\dots,(\vc{x}_n,y_n)\}$, the mini-batch size $b$ and the regularization coefficient $\lambda$, we propose to use Algorithm~\ref{alg:descending} to estimate the descent direction at the point $\omega_t$ for the $t^{th}$ iteration of deep transfer learning. 

With such descent direction estimator, the learning algorithm can replace the original stochastic gradient estimators used in stochastic gradient descent (SGD), Momentum and/or Adam for deep learning. Specifically, for each (e.g., the $t^{th}$) iteration of learning procedure, \TheName\ estimates the gradients of empirical loss and regularization term (i.e., $\nabla \widehat J_t$ and $\nabla \widehat\Omega_t$) separately. When the angle between gradients of empirical loss and regularization term is acute i.e., $\measuredangle(\nabla \widehat J_t,\nabla \widehat\Omega_t)\leq 90$, \TheName\ uses the original stochastic gradient as the descent direction (such as line 8 in Algorithm~\ref{alg:descending}). In such case, we believed the affect of regularization might not hurt the empirical loss minimization. On the other hand, when the angle is obtuse, \TheName\ decomposes the gradient of regularization term $\nabla\widehat{\Omega}_t$ to two orthogonal directions $\nabla\widehat{\Omega}^x_t$ and $\nabla\widehat{\Omega}^y_t$ and make $\nabla\widehat{\Omega}^x_t$ parallels with $\nabla \widehat{J}_t$, such that
\begin{equation}
    \nabla {\widehat \Omega^x_t} = \frac{< \nabla \hat{J}_t , \nabla \hat{\Omega}_t >}{||\nabla \hat{J}_t ||_2^2} \cdot \nabla \hat{J}_t,
\end{equation}
\begin{equation}
    \nabla {\widehat \Omega^y_t} =\nabla \hat{\Omega}_t - \nabla {\widehat \Omega^x_t}.
\end{equation}
Then \TheName\ truncates the direction against the gradient of empirical loss i.e., $\nabla\widehat{\Omega}^x_t$, and further compose the remaining direction $\nabla\widehat{\Omega}^y_t$ with gradient of empirical loss as the actual descent direction i.e., $\mathbf{\widehat d_t}=\lambda \widehat{J}_t+\lambda\cdot\nabla\widehat{\Omega}^y_t$.

\begin{algorithm}[H]
   
    \caption{\TheName:Descent Direction Estimation}
    \begin{algorithmic}[1]
    \Procedure{\TheName}{$\mathbf{D},\mathbf{\omega_t}, b,\lambda$}
     \State \textcolor{blue}{/*Stochastic Gradients Estimations*/}
     \State $\mathbf{B}_t\sim\mathbf{D}$ sampling a mini-batch of $b$ random samples from the training dataset
     \State $\nabla \widehat J_t$ estimating stochastic gradient of $J(\mathbf{\omega})$ at the point $\mathbf{\omega_t}$ using the mini-batch $\mathbf{B}_t$
     \State $\nabla \widehat \Omega_t$ estimating stochastic gradient of $\Omega_t(\mathbf{\omega},\mathbf{\omega_s})$ at the point $\mathbf{\omega_t}$  using the mini-batch $\mathbf{B}_t$
          \State \textcolor{blue}{/*Descent Direction Correction*/}
     \If{$\measuredangle(\nabla \widehat J_t,\nabla \widehat\Omega_t)\leq 90$}
     \State $\mathbf{\widehat d_t}\gets \nabla J_t+\lambda\cdot\nabla \widehat\Omega_t$
     \Else 
     \State $\nabla {\widehat \Omega^x_t}\perp\nabla {\widehat \Omega^y_t}\gets$ Orthogonal decomposition of $\nabla \widehat \Omega_t$ having $\nabla {\widehat \Omega^x_t}$ in parallel with $J_t$
     \State $\mathbf{\widehat d_t}\gets \nabla J_t+\lambda\cdot\nabla {\widehat \Omega^y_t}$
     \EndIf
     \State \Return{$\mathbf{\widehat d_t}$}
    \EndProcedure
    \end{algorithmic}
    \label{alg:descending}
  \end{algorithm}

\begin{figure}
    \centering
    \includegraphics[width=0.5\textwidth]{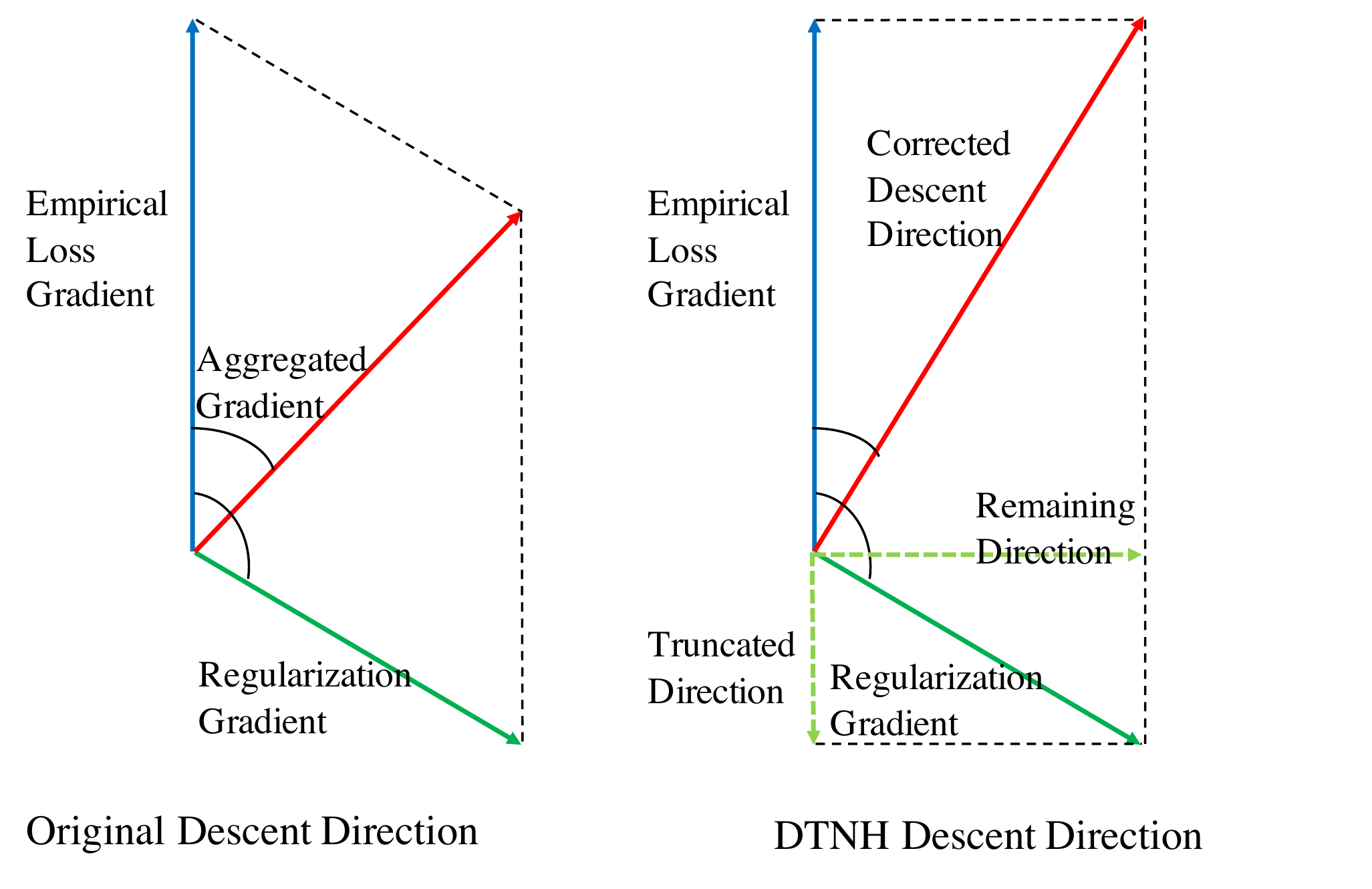}
    \caption{Example of \TheName\ Descent Direction Estimation}
    \label{fig:direction}
\end{figure}

For instance, Figure~\ref{fig:direction} illustrates an example of \TheName\ descent direction estimation, when the angles between gradients of empirical loss and regularization term is obtuse. The affect of regularization term forms a direction that might slow down the empirical loss descending.  \TheName\ decomposes the gradient of regularization term and truncates the conflicted direction for the actual descent direction estimation. On the other hand, the angle between the actual descent direction and regularization gradient is acute, so as to secure the regularization affect of knowledge transfer from pre-trained weights. 



\subsection{Discussion} 

Note that \TheName\ strategy is derived from the common stochastic gradient estimation used in stochastic gradient based learning algorithms, such as SGD, Momentum, conditioned SGD, Adam and so on. It can be consider as an alternative approach for descent direction estimation on top of vanilla stochastic gradient estimation, where you can still use natural gradient-alike method to condition the descent direction or adopt Momentum-alike acceleration methods to replace the weight updating mechanism.
We are not intending to compare \TheName\ with ANY gradient-based learning algorithms, as the contributions are complementary. One can freely use \TheName\ to improve any gradient-based optimization algorithms (if applicable) with the descend direction corrected.
\section{Experiment}
In this section, we reported our experiment results for \TheName. As was stated, we evaluated \TheName\ with the two types of regularization-based deep transfer learning paradigms, e.g., L2-SP \cite{li2018explicit} and Knowledge distillation based transfer~\cite{yim2017gift}.

\subsection{Data Sets and Experiment Setups}
Specifically, we used the ResNet-18~\cite{he2016deep} as our base model with three common source datasets including ImageNet \cite{deng2009imagenet},  Places 365~\cite{zhou2017places}, and Stanford Dogs 120~\cite{khosla2011novel} for weights pre-training. To demonstrate the performance of transfer learning, we further select five target datasets, including Caltech 256~\cite{griffin2007caltech}, MIT Indoors 67~\cite{quattoni2009recognizing}, Flowers 102~\cite{Nilsback08} and CIFAR 10~\cite{krizhevsky2014cifar}. Note that, we follow the same settings used in~\cite{li2018explicit} for Caltech 256 setup, where 30 or 60 samples randomly drawn from each category for training with 20 remaining samples for testing. Table~\ref{tab:stats} presents the statistics on some basic facts of the 7 datasets used in this experiments. 

\begin{table}[]
\centering
            \caption{Statistics on Source/Target Datasets}
\normalsize{
\begin{tabular}{c||c c}\hline
        Datasets &  Domains & \# Train/Test \\\hline 
        \multicolumn{3}{c}{Source Tasks}\\\hline
        ImageNet &  visual objects  & 1,419K+/100K\\
        Place 365 & indoor scenes & 10,000K+\\
        Stanford Dog 120 & dogs & 12K/8.5K\\\hline
        \multicolumn{3}{c}{Target Tasks}\\\hline
        CIFAR 10  & objects & 50K/10K\\
        Caltech 256 & objects & 30K+\\
        MIT Indoors 67 & indoor scenes & 5K+/1K+\\
        Flowers 102  & flowers & 1K+/6K+\\
       \hline
    \end{tabular}}
    \label{tab:stats}    
\end{table}

Furthermore, to obtain the pre-trained weights of all source tasks, we adopt the pre-trained models of ImageNet~\footnote{https://github.com/PaddlePaddle/models}, Place 365~\footnote{https://github.com/CSAILVision/places365}, and Stanford Dog 120\footnote{https://github.com/stormy-ua/dog-breeds-classification} released online. We found an interesting fact that the pre-trained models of Place 365 and Stanford Dog 120 were trained from the pre-trained model of ImageNet. In this way, the pre-trained models for Place 365 and Stanford Dog 120 have been already enhanced by the ImageNet.

\textbf{Source/Target Tasks Pairing} Above configuration leads to 15 source/target task pairs, where regularization would hurt the performance of transfer learning in some of these cases. For example, the image contents of ImageNet and CIFAR 10 are quite similar, in this way, the knowledge transfer from ImageNet to CIFAR 10 could improve the performance. On the other hand, the images in Stanford Dog 120 and  MIT Indoor 67 are quite different, e.g., dogs v.s. indoor scenes; then the regularization based on pre-trained weights of Stanford Dog 120 task would hurt the learning of MIT Indoor 67 task.

   
\textbf{Image Classification Tasks Setups.} All images are re-sized to $256 \times 256$ and normalized to zero mean for each channel, following with data augmentation operations of random mirror and random crop to $224 \times 224$. We use a batch size of $48$, SGD with the momentum of 0.9 is used for optimizing all models~\cite{sutskever2013importance}. The learning rate for base model starts with 0.01 and is divided by 10 after 6000 iterations. The Training is terminated with 8000 iterations for Caltech 256, MIT Indoor 67 and Flowers 102, terminates with 20,000 iterations for CIFAR 10 (i.e., 18 epochs). Cross validation has been done for searching the best regularization coefficient. Note that regularization coefficient decays at certain ratio, ranging from 0 to 1, per epoch. The pre-trained weights obtained from the source task were not only used as the initialization of the model, i.e., starting point of optimization. Under the best configuration, each experiment is repeated five times. The averaged accuracy with standard deviations are reported in this paper.

\textbf{Hyper-parameter Tuning.} The tuning parameters (i.e., $\lambda$ in Eq.~\ref{eq:deep})  for all experiments have been tuned best using cross validation. Please see also in the top-1 accuracy for the deep transfer learning algorithms listed in Tables~II,~III and~IV. We reproduced the experiments using Fine-tuning, $L^2$-SP~\cite{li2018explicit} and KnowDist~\cite{yim2017gift} algorithms based on the benchmark datasets, where we can find that these baseline algorithms indeed achieved better accuracy than the original work~\cite{li2018explicit,yim2017gift} under the same settings. 


\begin{table*}[!h]
    \centering
    \caption{Classification Accuracy Comparison from Source \textbf{ImageNet} (Numbers in \textcolor{red}{RED} refer to the evidence of possible negative transfer)}
\normalsize{
    \begin{tabular}{c||c|cc|cc}
    \hline
  Target Datasets & Fine-tuning & $L^2$-SP & \TheName\ ($L^2$-SP) & KnowDist &\TheName\ (KnowDist) \\
    \hline
    Caltech 256  & $82.68 \pm 0.2$ & $83.69 \pm 0.09$ & $\bm{84.14 \pm 0.08}$ & $82.93 \pm 0.08$ & $\bm{83.27 \pm 0.4}$\\
     MIT Indoors 67 & $76.73 \pm 0.77$ & \textcolor{red}{$ 75.11 \pm 0.43 $} & $\bm{77.46 \pm 0.29}$ & $78.05 \pm 0.32$ & $\bm{78.77 \pm 0.31}$\\
     Flowers 102 & $90.24 \pm 0.31$ & \textcolor{red}{$88.96 \pm 0.21$} & $\bm{90.68 \pm 0.31}$ & $90.43 \pm 0.4$& $\bm{90.91 \pm 0.4}$\\
     CIFAR 10 &  $96.40\pm 0.4$ & \textcolor{red}{$93.30 \pm 0.16$} & $\bm{96.41 \pm 0.11}$ & $96.43 \pm 0.08$ & $\bm{96.57 \pm 0.2}$ \\
    \hline
    \end{tabular}}
    \label{tab:1}
\end{table*}

\begin{table*}[!h]
    \centering
    \caption{Classification Accuracy Comparison from Source \textbf{Places 365} (Numbers in \textcolor{red}{RED} refer to the evidence of possible negative transfer)}
\normalsize{
    \begin{tabular}{c||c|cc|cc}
    \hline
    Target Databsets & Fine-tuning & $L^2$-SP & \TheName\ ($L^2$-SP) & KnowDist &\TheName\ (KnowDist) \\
    \hline
     Caltech 256 & $73.13 \pm 0.20$ & \textcolor{red}{$66.99 \pm 0.20$} & $\bm{73.32 \pm 0.1}$ & \textcolor{red}{$72.8 \pm 0.22$} & $\bm{73.18 \pm 0.24}$\\
     MIT Indoors 67 & $82.64 \pm 0.16$ & $84.09 \pm 0.09$ & $\bm{84.19 \pm 0.07}$ & $83.29 \pm 0.42$ & $\bm{84.40 \pm 0.41}$\\ 
     Flowers 102 & $83.77 \pm 0.68$ & \textcolor{red}{$77.66 \pm 0.13$} & $\bm{84.11 \pm 0.06}$ & \textcolor{red}{$83.50 \pm 0.26$} & $\bm{84.12 \pm 0.56}$\\
     CIFAR 10 & $89.35 \pm 0.59$ & $89.78 \pm 0.05$ & $\bm{90.85 \pm 0.11}$ & $94.96 \pm 0.05$ & $\bm{95.02 \pm 0.13}$\\
    \hline
    \end{tabular}}
    \label{tab:2}
\end{table*}

\begin{table*}[!h]
    \centering
        \caption{Classification Accuracy Comparison from Source \textbf{Stanford Dogs 120} (Numbers in \textcolor{red}{RED} refer to the evidence of possible negative transfer)}
\normalsize{
    \begin{tabular}{c||c|cc||cc}
    \hline
    Target Datasets & Fine-tuning & $L^2$-SP & \TheName\ ($L^2$-SP) & KnowDist &\TheName\ (KnowDist) \\
    \hline
     Caltech 256  & $82.29 \pm 0.04$ & $83.44 \pm 0.23$ & $\bm{83.84 \pm 0.08}$ & $82.73 \pm 0.26$ & $\bm{82.85 \pm 0.27}$\\
     MIT Indoors 67  & $75.69 \pm 0.21$ & \textcolor{red}{$74.64 \pm 0.07$} & $\bm{76.46 \pm 0.22}$ & $76.36 \pm 0.19$ & $\bm{76.74 \pm 026}$\\ 
     Flowers 102  & $90.20 \pm 0.39$ & \textcolor{red}{$88.14 \pm 0.06$} & $\bm{89.98 \pm 0.04}$ & \textcolor{red}{$89.86 \pm 0.07$} & $\bm{90.29 \pm 0.34}$\\
     CIFAR 10  & $96.34 \pm 0.13 $ &\textcolor{red}{$94.16 \pm 0.10$} & $\bm{96.39 \pm 0.08}$ & \textcolor{red}{$96.11 \pm 0.53$} & $\bm{96.41 \pm 0.18}$\\
    \hline
    \end{tabular}}
    \label{tab:3}
\end{table*}

\subsection{Overall Performance Comparison}
In this section, we report the results of overall performance comparison based on the above 15 source/target tasks pairs using two deep transfer learning regularizers---$L^2$-SP~\cite{li2018explicit} and Knowledge Distillation (namely {KnowDist})~\cite{yim2017gift}. Here, we mainly focus on evaluate the performance improvement contributed by \TheName\ on top of $L^2$-SP and KnowDist, comparing to the vanilla implementations of these two algorithms. We evaluate the four algorithms---\TheName\ based on $L^2$-SP, \TheName\ based on KnowDist, the vanilla implementations of $L^2$-SP and KnowDist, using the all aforementioned 15 pairs of source/target tasks, under the same machine learning settings. We present the overall accuracy comparisons in Tables~\ref{tab:1},~\ref{tab:2}, and~\ref{tab:3}.
%
%
It is obvious that \TheName\ can significantly improve  $L^2$-SP and KnowDist to achieve better accuracy in all the 15 source/target pairs. 

\subsection{General Transfer Learning Cases}
\TheName\ improves the performance of deep transfer learning in above cases, no matter, whether negative transfer occurs. For example, CIFAR 10 target task works well with source task ImageNet using $L^2$-SP algorithm achieving 93.30\% accuracy, while \TheName\ ($L^2$-SP) can improve it under the same setting with accuracy 96.41\% (with more than 3.1\% accuracy improvement). For the same experiment, KnowDist can achieve 96.43\%, while \TheName\ (KnowDist) can further improve it to 96.57\%. To the best of our knowledge, it might be the known limit~\cite{recht2018cifar} for CIFAR 10 training from ImageNet sources with only 18 epochs.

An interesting facts observed in the experiments is that, on top of the all four algorithms and 15 source/task pairs, using Stanford Dog 120 as the source task can perform similar as the ones sourcing from ImageNet. We consider it is due to the reason that the public release of Stanford Dog 120 pre-trained model is pre-trained from ImageNet, while the size of Stanford Dog 120 dataset is relatively small (i.e., it cannot ``wash out'' the knowledge obtained from ImageNet while preserving the knowledge from the both ImageNet/Stanford Dog 120 datasets). In this way, knowledge transferring from Stanford Dog 120 can be as good as those based on ImageNet. In the meanwhile, \TheName\ can still improve the performance of $L^2$-SP and KnowDist, gaining 0.12\%$\sim$2.2\% higher accuracy with low variance, even given the well-trained Stanford Dog 120 model. 

\subsubsection{Performance with Negative Transfer Effect} 
According to the results presented in Tables~\ref{tab:1},~\ref{tab:2}, and~\ref{tab:3}, we find negative transfer may happen in the cross-domain cases ``Visual Objects/Dogs $\Leftrightarrow$ Indoor Scenes'' (please refer to the domain definitions in Table~\ref{tab:stats}), while \TheName\ can improve the performance of $L^2$-SP and KnowDist to relieve such negative effects. Two detailed cases are addressed as follow.

\textbf{Cases of Negative Transfer} For both $L^2$-SP and KnowDist algorithms, when using ImageNet and Stanford Dogs 120 as the source task while transferring to MIT Indoors 67 as the target task, we can observe significant performance degradation comparing to knowledge transfer from Place 365 to MIT Indoor 67. For example (\textbf{Case I}), the accuracy of MIT Indoor 67 using $L^2$-SP is 84.09\% based on pre-trained weights of Place 365, while the accuracy would be degraded to 75.11\% and 74.64\% under the same settings with ImageNet and Stanford Dog 120 as the pre-trained models respectively. Furthermore, we also observe the similar negative transfer effects, when using Place 365 as source while transfer to the target tasks based on Caltech 256, Flower 102 and CIFAR 10. For example (\textbf{Case II}), the accuracy on Flowers 102 is 77.66\% using Place 365 as source, while sourcing from ImageNet and Stanford Dog can achieve as high as 88.96\% and 88.14\% respectively, all based on $L^2$-SP. 

\textbf{Relieving Negative Transfer Effects.} We believe performance degradation appeared in \textbf{Cases I} and \textbf{II} is due to the negative transfer, as the domains of these datasets are quite different. \TheName\ can however relieve such negative transfer cases. \TheName\ ($L^2$-SP) can achieve 84.11\% on Flowers 102 dataset even when sourcing from Place 365 --- i.e., achieving 7\% accuracy improvement, comparing to vanilla $L^2$-SP under the same settings. For the rest negative transfer cases, \TheName\ can still improve the performance, with around 2\% higher accuracy, comparing to the vanilla implementations of $L^2$-SP and KnowDist algorithms. In this way, we conclude that \TheName\ can improve the performance of $L^2$-SP and KnowDist in negative transfer cases with $2\%\sim 7\%$ higher accuracy. 

Note that we don't intend to claim \TheName\ eliminating the negative transfer effects. It, however, improves the performance of regularization-based deep transfer learning, even with inappropriate source/target pairs. Such accuracy improvement can marginally solve the problem of negative transfer effects.

\subsection{Case Studies}
We plan to report the results of following two cases studies that directly prove \TheName\ working in the way we assumed.

\subsubsection{Empirical Loss Minimization}
As was elaborated in introduction section, we doubt that using regularizer might restrict the learning procedure from lowering the empirical loss of deep transfer learning. Such restriction helps the deep transfer learning to avoid over-fitting, but in the meanwhile, hurts the learning procedure. In this way, we hope to study trends of empirical loss part minimization with and without \TheName\ using the regularization-base deep transfer learning algorithms. Note that the empirical loss here is NOT the training loss, it refers to the data fitting error part of the training loss.

\begin{figure}[!h]
    \centering
    \includegraphics[width=0.5\textwidth]{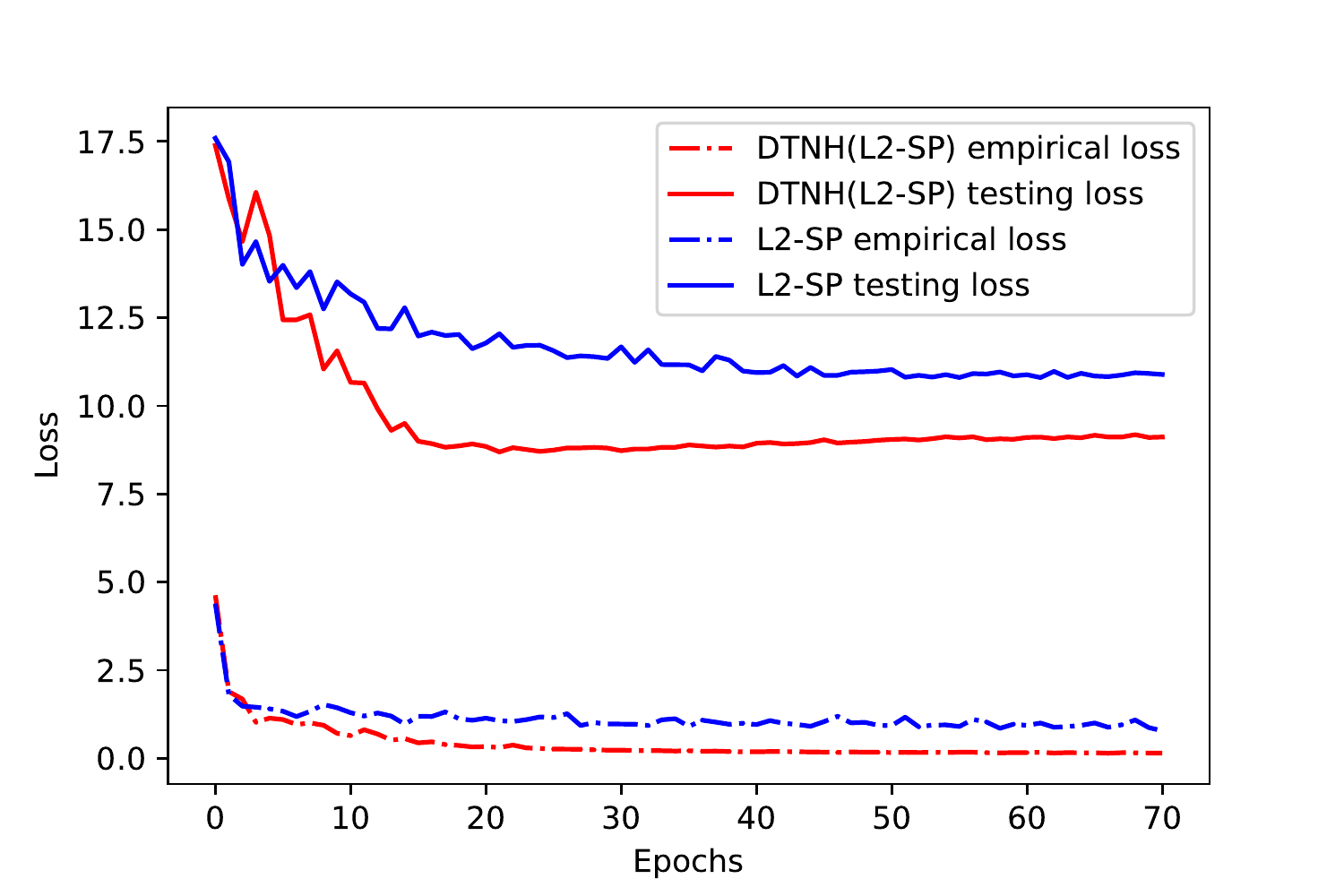}
    \caption{Empirical Loss Minimization}
    \label{fig:loss}
\end{figure}

Figure~\ref{fig:loss} illustrates the trends of both empirical loss and testing loss, with increasing number of iterations, based on both $L^2$-SP and  \TheName\ ($L^2$-SP), for Places 365 $\Leftrightarrow$ MIT Indoors 67 case. As was expected, the empirical loss of both vanilla $L^2$-SP and \TheName\ ($L^2$-SP) reduces with the number of iterations, while the empirical loss of $L^2$-SP is always higher than that of \TheName ($L^2$-SP).
In the meanwhile, \TheName ($L^2$-SP) always enjoys a lower testing loss than vanilla $L^2$-SP. 
The phenomena indicates that, compared vanilla $L^2$-SP to \TheName ($L^2$-SP), the $L^2$-SP regularization term would restrict the procedure of empirical loss minimization and finally hurt the learning procedure with lower testing accuracy.


\subsubsection{Descent Direction vs Original Gradients}
The intuition of \TheName\ design is based on our two assumptions made in section 3.1--- it is possible to find a new descent direction that can be very closed to the direction of empirical loss gradient (\textbf{Assumption 1}), while always sharing a small angle with the gradient of regularization term (\textbf{Assumption 2}).

In this case, we hope to see the angle(denoted as {\bf Angle 1}) between the actual descent direction of \TheName ($L^2$-SP) and the (stochastic) gradient of empirical loss, i.e., a noisy estimation of $\measuredangle (\mathbf{\widehat d}(\mathbf{\omega}),\nabla J(\mathbf{\omega}))$ (defined in section 3.1), and the angle(denoted as {\bf Angle 2}) between the actual descent direction of \TheName ($L^2$-SP) and the (stochastic) gradient of $L^2$-SP regularization term, i.e., a noisy estimation of $\measuredangle (\mathbf{\widehat d}(\mathbf{\omega}),\nabla \Omega(\mathbf{\omega},\mathbf{\omega}_s))$ (defined in section 3.1). 

\begin{figure}
   \centering
    \includegraphics[width=0.5\textwidth]{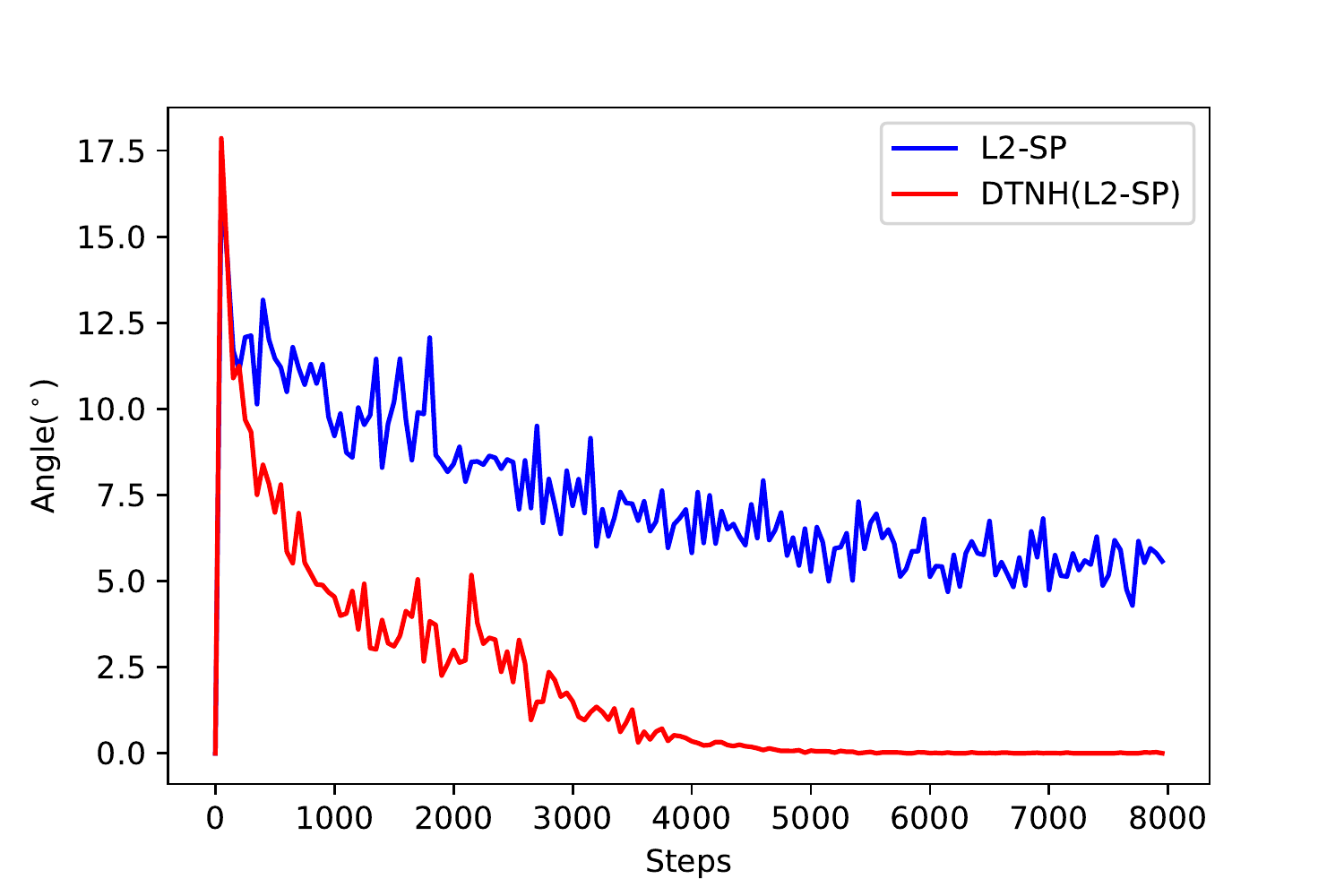}
    \caption{Angle 1 vs Angle 3}
    \label{fig:angle2}
\end{figure}

\begin{figure}[!h]
    \centering
    \includegraphics[width=0.5\textwidth]{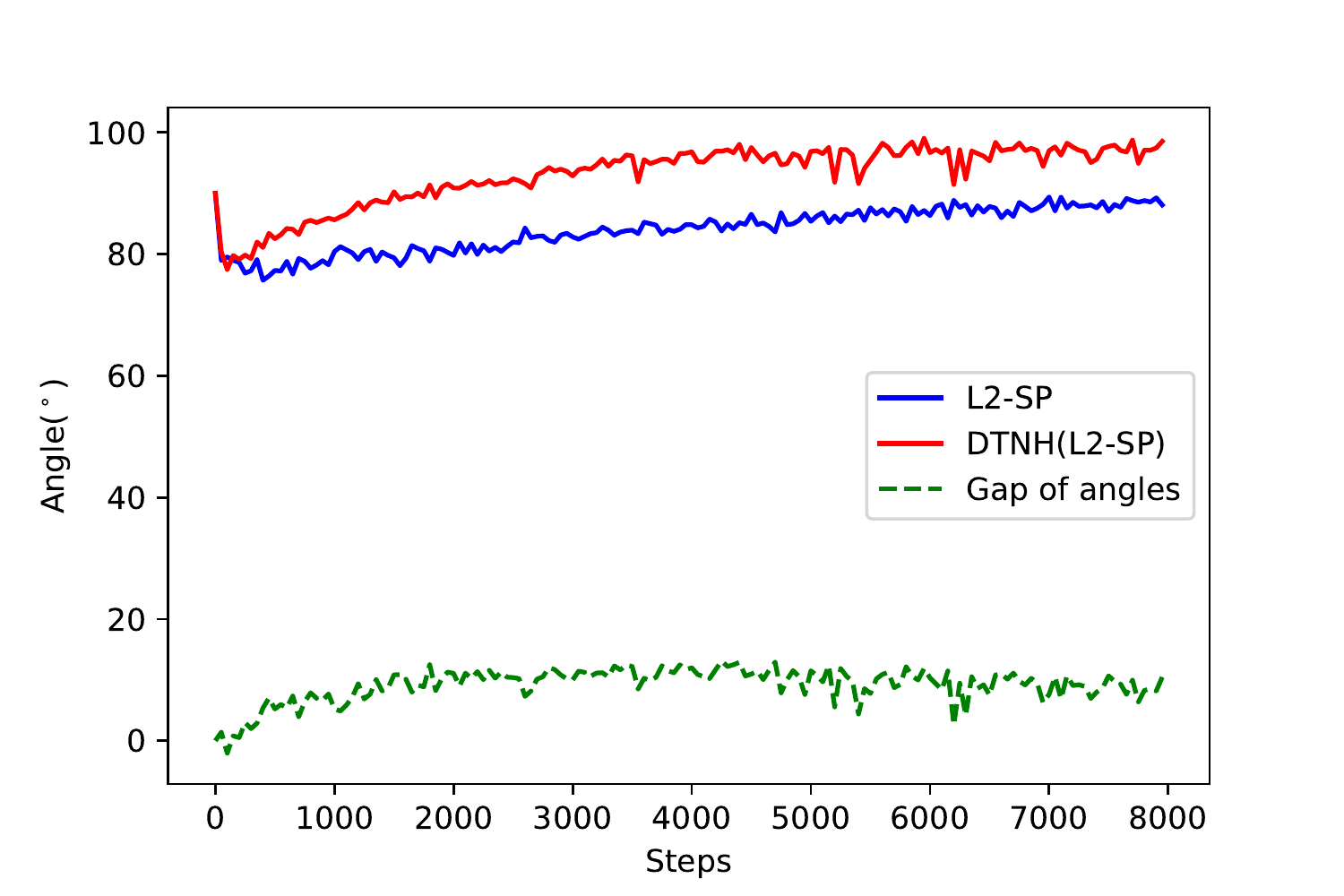}
    \caption{Angle 2 vs. Angle 4}
   \label{fig:angle3}
\end{figure}

\begin{enumerate}
  
  \item \emph{Validation on Assumption 1.} As shown in Figure~\ref{fig:angle2}, we may compare {\bf Angle 1} with the angle(denoted as {\bf Angle 3}) between the (stochastic) gradient of vanilla $L^2$-SP vs. the (stochastic gradient) of empirical loss, i.e., a noisy estimation of $\measuredangle (\nabla \mathcal{L}(\mathbf{\omega}),\nabla J(\mathbf{\omega}))$. As was discussed in section 3.1, when {\bf Angle 1} is smaller than {\bf Angle 3}, then we can say \TheName\ is on the direction that minimizes the empirical loss faster than $L^2$-SP.

  \item \emph{Validation on Assumption 2.} As shown in Figure~\ref{fig:angle3}, we would further compare {\bf Angle 2} with the angle(denoted as {\bf Angle 4}) between the (stochastic) gradient of vanilla $L^2$-SP and the (stochastic gradient) of $L^2$-SP regularization term, i.e., a noisy estimation of $\measuredangle (\nabla \mathcal{L}(\mathbf{\omega}),\nabla \Omega(\mathbf{\omega},\mathbf{\omega}_s))$ . When {\bf Angles 2} and {\bf 4} with a small gap, then we can say the {\bf Angle 2} is relatively small. In this case, \TheName\ shares a descent direction affecting by the regularizer, it still preserves the power of regularizer. 
\end{enumerate}
%


\section{Discussions}
In this paper, we claimed one of our major contribution is to alleviate the ``negative transfer'' phenomenon caused by the reuse of inappropriate pre-trained weights for regularization-based transfer learning (e.g., $L^2-SP$~\cite{li2018explicit} and KnowDist~\cite{yim2017gift}). 

In terms of evidence, some previous work~\cite{rosenstein2005transfer,kandaswamy2014improving,weiss2016survey} has already demonstrated the effects of ``negative transfer''. In addition, our experiment also provided 15 detailed real-life cases: using the pre-trained weights from 3 sources datasets (i.e., ImageNet,  Places 365 and Stanford Dogs 120) and transferring to 5 different target tasks (i.e., Caltech 30, Caltech 60, MIT Indoors 67, Flowers 102, and CIFAR 10), where we present the comparison results in Tables~II,~III, and IV in the Section V. It has been shown that, while regularization-based transfer learning can outperform fine-tuning in most cases, the performance of regularization performed even worse than fine-tuning for some specific source-target pairs.  Our experiment results validate the existence of ``negative transfer'' effects. It also suggests one can alleviate the ``negative transfer'' effects that used to appear in {``inappropriate source-target pairs''} through incorporating the proposed algorithm \TheName, and achieve better accuracy than fine-tuning and the vanilla implementation of $L^2$-SP~\cite{li2018explicit} and KnowDist~\cite{yim2017gift}.

To evaluate the improvement of \TheName\ beyond the regularization-based deep transfer learning algorithms, we use $L^2$-SP~\cite{li2018explicit} and KnowDist~\cite{yim2017gift} as the reference. To our knowledge, $L^2$-SP~\cite{li2018explicit} and KnowDist~\cite{yim2017gift} are considered as the state-of-the-art transfer learning algorithms without any modifications to the deep architectures. A new state of the art algorithm for regularization-based deep transfer learning is DELTA~\cite{li2019delta} that uses feature maps over attention mechanisms as regularization to further improve KnowDist~\cite{yim2017gift}. In our experiment, we did not include DELTA for the comparison of baselines. Since both DELTA and KnowDist use the regularization between the feature maps to enable the knowledge transfer from teacher to student networks, we thus believe \TheName\ should work with the algorithms like DELTA.

In terms of methodology, the most relevant work to our study is Gradient Episodic Memory (GEM) for continual learning~\cite{lopez2017gradient}, which continuously learn the new task using the well-trained models for past tasks through regularization terms. In terms of objectives, \TheName\ aims at lowering the effects of knowledge transfer regularization from hurting empirical loss minimization, while GEM prevents the empirical loss minimization from hurting regularization effects~(i.e., the accuracy on old tasks). 
In terms of algorithms, in every iteration of learning, GEM estimates the descend direction with respect to the gradients of the new task and all past tasks using a time-consuming Quadratic Program~(QP), while \TheName\ re-estimates the descend direction from the gradients of the regularizer term and the empirical loss term with low-complexity orthogonal decomposition. All in all, GEM can be considered as a special case of \TheName\ using L$^2$-SP regularizer~\cite{li2018explicit} based on two tasks.

\section{Conclusions}
In this paper, we studied a descent direction estimation strategy \TheName\ that improves the common regularization techniques, such as $L^2$-SP~\cite{li2018explicit} and Knowledge Distillation~\cite{yim2017gift}, for deep transfer learning. Non-trivial contribution has been made compared to the existing methods that simple aggregates empirical loss for data fitting and regularizer for knowledge transfer through linear combination, such as~\cite{li2018explicit,yim2017gift}.

Specifically, we designed a new method to re-estimate a new direction for loss descending based on the (stochastic) gradient estimation of empirical loss and regularizers, where orthogonal decomposition has been made on the gradient of regularizers, so as to eliminate the conflicted direction against the empirical loss descending. We conduct extensive experiments to evaluate \TheName\ using several real-world datasets based on typical convolutional neural networks. The experiment results and comparisons show that \TheName\ can significantly outperform the state of the arts with higher accuracy, even in the negative transfer cases.

\section*{Acknowledgement}
We would appreciate the program committee and reviewers' efforts in reviewing and improving the manuscript. This paper was done when Mr. Ruosi Wan was a full-time research intern at Baidu Inc. Please refer to the open-source repository for the PaddlePaddle based implementation of \TheName. The deep transfer learning algorithms proposed in this paper have been transferred to technologies adopted by PaddleHub\footnote{https://github.com/PaddlePaddle/PaddleHub} and Baidu EZDL\footnote{https://ai.baidu.com/ezdl/}. The first two authors contributed equally to this paper. While Mr. Ruosi Wan contributed the algorithm design, Dr. Haoyi Xiong led the research and wrote parts of the paper. Please contact Dr. Haoyi Xiong via xionghaoyi@baidu.com for correspondence. 
{
\bibliographystyle{ieee}
\bibliography{main}

\begin{thebibliography}{10}\itemsep=-1pt

\bibitem{aygun2017exploiting}
M.~Aygun, Y.~Aytar, and H.~K. Ekenel.
\newblock Exploiting convolution filter patterns for transfer learning.
\newblock In {\em ICCV Workshops}, pages 2674--2680, 2017.

\bibitem{caruana1997multitask}
R.~Caruana.
\newblock Multitask learning.
\newblock {\em Machine learning}, 28(1):41--75, 1997.

\bibitem{cui2018large}
Y.~Cui, Y.~Song, C.~Sun, A.~Howard, and S.~Belongie.
\newblock Large scale fine-grained categorization and domain-specific transfer
  learning.
\newblock In {\em Proceedings of the IEEE Conference on Computer Vision and
  Pattern Recognition}, pages 4109--4118, 2018.

\bibitem{deng2009imagenet}
J.~Deng, W.~Dong, R.~Socher, L.-J. Li, K.~Li, and L.~Fei-Fei.
\newblock Imagenet: A large-scale hierarchical image database.
\newblock In {\em Computer Vision and Pattern Recognition, 2009. CVPR 2009.
  IEEE Conference on}, pages 248--255. Ieee, 2009.

\bibitem{donahue2014decaf}
J.~Donahue, Y.~Jia, O.~Vinyals, J.~Hoffman, N.~Zhang, E.~Tzeng, and T.~Darrell.
\newblock Decaf: A deep convolutional activation feature for generic visual
  recognition.
\newblock In {\em International conference on machine learning}, pages
  647--655, 2014.

\bibitem{ge2017cvpr}
W.~Ge and Y.~Yu.
\newblock Borrowing treasures from the wealthy: Deep transfer learning through
  selective joint fine-tuning.
\newblock In {\em Proceedings of the IEEE conference on computer vision and
  pattern recognition}, pages 10--19, 2017.

\bibitem{griffin2007caltech}
G.~Griffin, A.~Holub, and P.~Perona.
\newblock Caltech-256 object category dataset.
\newblock 2007.

\bibitem{he2016deep}
K.~he, X.~Zhang, S.~Ren, and J.~Sun.
\newblock Deep residual learning for image recognition.
\newblock In {\em Proceedings of the IEEE conference on computer vision and
  pattern recognition}, pages 770--778, 2016.

\bibitem{hinton2015distilling}
G.~Hinton, O.~Vinyals, and J.~Dean.
\newblock Distilling the knowledge in a neural network.
\newblock {\em arXiv preprint arXiv:1503.02531}, 2015.

\bibitem{huh2016makes}
M.~Huh, P.~Agrawal, and A.~A. Efros.
\newblock What makes imagenet good for transfer learning?
\newblock {\em arXiv preprint arXiv:1608.08614}, 2016.

\bibitem{kandaswamy2014improving}
C.~Kandaswamy, L.~M. Silva, L.~A. Alexandre, J.~M. Santos, and J.~M. de~S{\'a}.
\newblock Improving deep neural network performance by reusing features trained
  with transductive transference.
\newblock In {\em International Conference on Artificial Neural Networks},
  pages 265--272. Springer, 2014.

\bibitem{khosla2011novel}
A.~Khosla, N.~Jayadevaprakash, B.~Yao, and F.-F. Li.
\newblock Novel dataset for fine-grained image categorization: Stanford dogs.
\newblock In {\em Proc. CVPR Workshop on Fine-Grained Visual Categorization
  (FGVC)}, volume~2, page~1, 2011.

\bibitem{kirkpatrick2017overcoming}
J.~Kirkpatrick, R.~Pascanu, N.~Rabinowitz, J.~Veness, G.~Desjardins, A.~A.
  Rusu, K.~Milan, J.~Quan, T.~Ramalho, A.~Grabska-Barwinska, et~al.
\newblock Overcoming catastrophic forgetting in neural networks.
\newblock {\em Proceedings of the national academy of sciences}, page
  201611835, 2017.

\bibitem{krizhevsky2014cifar}
A.~Krizhevsky, V.~Nair, and G.~Hinton.
\newblock The cifar-10 dataset.
\newblock {\em online: http://www. cs. toronto. edu/kriz/cifar. html}, 2014.

\bibitem{li2018explicit}
X.~Li, Y.~Grandvalet, and F.~Davoine.
\newblock Explicit inductive bias for transfer learning with convolutional
  networks.
\newblock {\em Thirty-fifth International Conference on Machine Learning},
  2018.

\bibitem{li2019delta}
X.~Li, H.~Xiong, H.~Wang, Y.~Rao, L.~Liu, and J.~Huan.
\newblock {DELTA}: {DEEP} {LEARNING} {TRANSFER} {USING} {FEATURE} {MAP} {WITH}
  {ATTENTION} {FOR} {CONVOLUTIONAL} {NETWORKS}.
\newblock In {\em International Conference on Learning Representations}, 2019.

\bibitem{li2017learning}
Z.~Li and D.~Hoiem.
\newblock Learning without forgetting.
\newblock {\em IEEE Transactions on Pattern Analysis and Machine Intelligence},
  2017.

\bibitem{liu2017sparse}
J.~Liu, Y.~Wang, and Y.~Qiao.
\newblock Sparse deep transfer learning for convolutional neural network.
\newblock In {\em AAAI}, pages 2245--2251, 2017.

\bibitem{lopez2017gradient}
D.~Lopez-Paz et~al.
\newblock Gradient episodic memory for continual learning.
\newblock In {\em Advances in Neural Information Processing Systems}, pages
  6467--6476, 2017.

\bibitem{mnih2014recurrent}
V.~Mnih, N.~Heess, A.~Graves, et~al.
\newblock Recurrent models of visual attention.
\newblock In {\em Advances in neural information processing systems}, pages
  2204--2212, 2014.

\bibitem{Nilsback08}
M.-E. Nilsback and A.~Zisserman.
\newblock Automated flower classification over a large number of classes.
\newblock In {\em Proceedings of the Indian Conference on Computer Vision,
  Graphics and Image Processing}, Dec 2008.

\bibitem{pan2010survey}
S.~J. Pan, Q.~Yang, et~al.
\newblock A survey on transfer learning.
\newblock {\em IEEE Transactions on knowledge and data engineering},
  22(10):1345--1359, 2010.

\bibitem{quattoni2009recognizing}
A.~Quattoni and A.~Torralba.
\newblock Recognizing indoor scenes.
\newblock In {\em 2009 IEEE Conference on Computer Vision and Pattern
  Recognition}, pages 413--420. IEEE, 2009.

\bibitem{recht2018cifar}
B.~Recht, R.~Roelofs, L.~Schmidt, and V.~Shankar.
\newblock Do cifar-10 classifiers generalize to cifar-10?
\newblock {\em arXiv preprint arXiv:1806.00451}, 2018.

\bibitem{romero2014fitnets}
A.~Romero, N.~Ballas, S.~E. Kahou, A.~Chassang, C.~Gatta, and Y.~Bengio.
\newblock Fitnets: Hints for thin deep nets.
\newblock {\em arXiv preprint arXiv:1412.6550}, 2014.

\bibitem{rosenstein2005transfer}
M.~T. Rosenstein, Z.~Marx, L.~P. Kaelbling, and T.~G. Dietterich.
\newblock To transfer or not to transfer.
\newblock In {\em NIPS 2005 workshop on transfer learning}, volume 898, pages
  1--4, 2005.

\bibitem{sutskever2013importance}
I.~Sutskever, J.~Martens, G.~Dahl, and G.~Hinton.
\newblock On the importance of initialization and momentum in deep learning.
\newblock In {\em International conference on machine learning}, pages
  1139--1147, 2013.

\bibitem{weiss2016survey}
K.~Weiss, T.~M. Khoshgoftaar, and D.~Wang.
\newblock A survey of transfer learning.
\newblock {\em Journal of Big Data}, 3(1):9, 2016.

\bibitem{xu2015show}
K.~Xu, J.~Ba, R.~Kiros, K.~Cho, A.~Courville, R.~Salakhudinov, R.~Zemel, and
  Y.~Bengio.
\newblock Show, attend and tell: Neural image caption generation with visual
  attention.
\newblock In {\em International conference on machine learning}, pages
  2048--2057, 2015.

\bibitem{yang2016stacked}
Z.~Yang, X.~He, J.~Gao, L.~Deng, and A.~Smola.
\newblock Stacked attention networks for image question answering.
\newblock In {\em Proceedings of the IEEE Conference on Computer Vision and
  Pattern Recognition}, pages 21--29, 2016.

\bibitem{yim2017gift}
J.~Yim, D.~Joo, J.~Bae, and J.~Kim.
\newblock A gift from knowledge distillation: Fast optimization, network
  minimization and transfer learning.
\newblock In {\em The IEEE Conference on Computer Vision and Pattern
  Recognition (CVPR)}, volume~2, 2017.

\bibitem{yosinski2014transferable}
J.~Yosinski, J.~Clune, Y.~Bengio, and H.~Lipson.
\newblock How transferable are features in deep neural networks?
\newblock In {\em Advances in neural information processing systems}, pages
  3320--3328, 2014.

\bibitem{zagoruyko2016paying}
S.~Zagoruyko and N.~Komodakis.
\newblock Paying more attention to attention: Improving the performance of
  convolutional neural networks via attention transfer.
\newblock {\em arXiv preprint arXiv:1612.03928}, 2016.

\bibitem{zhang2018parameter}
Y.~Zhang, Y.~Zhang, and Q.~Yang.
\newblock Parameter transfer unit for deep neural networks.
\newblock {\em arXiv preprint arXiv:1804.08613}, 2018.

\bibitem{zhou2017places}
B.~Zhou, A.~Lapedriza, A.~Khosla, A.~Oliva, and A.~Torralba.
\newblock Places: A 10 million image database for scene recognition.
\newblock {\em IEEE Transactions on Pattern Analysis and Machine Intelligence},
  2017.

\end{thebibliography}
}

\end{document}